\documentclass[11pt,a4paper]{article}

\usepackage[utf8]{inputenc}
\usepackage[T1]{fontenc}
\usepackage{amsmath,amssymb}
\usepackage{graphicx}
\usepackage{booktabs}
\usepackage{hyperref}
\usepackage[round]{natbib}
\usepackage{xcolor}
\usepackage{microtype}
\usepackage{geometry}
\title{Semantic Novelty Trajectories in 80,000 Books:\\A Cross-Corpus Embedding Analysis}
\author{W.\ Frederick Zimmerman\\
  Nimble Books LLC, Ann Arbor, MI\\
  \texttt{wfz@nimblebooks.com}}
\date{March 2026}

\begin{document}
\maketitle

\begin{abstract}
I apply Schmidhuber's compression progress theory of interestingness at
corpus scale, analyzing semantic novelty trajectories in more than 80,000
books spanning two centuries of English-language publishing.  Using
sentence-transformer paragraph embeddings and a running-centroid novelty
measure, I compare 28,730 pre-1920 Project Gutenberg books (PG19) against
52,796 modern English books (Books3, approximately 1990--2010).  The
principal findings are fourfold.  First, mean paragraph-level novelty is
roughly 10\% higher in modern books (0.503 vs.\ 0.459).  Second,
trajectory circuitousness---the ratio of cumulative path length to net
displacement in embedding space---nearly doubles in the modern corpus
($+67\%$).  Third, convergent narrative curves, in which novelty declines
toward a settled semantic register, are 2.3$\times$ more common in
pre-1920 literature.  Fourth, novelty is orthogonal to reader quality
ratings ($r = -0.002$), suggesting that interestingness in
Schmidhuber's sense is structurally independent of perceived literary
merit.  Clustering paragraph-level trajectories via PAA-16 representations
reveals eight distinct narrative-shape archetypes whose distribution shifts
substantially between eras.  All analysis code and an interactive
exploration toolkit are publicly available at
\url{https://bigfivekiller.online/novelty_hub}.
\end{abstract}


\section{Introduction}
\label{sec:introduction}

What makes a text \emph{interesting}?  \citet{schmidhuber2009} proposed a
formal answer grounded in algorithmic information theory: an observer finds
a data stream interesting to the extent that a learning algorithm can
compress it better over time.  The momentary ``interestingness'' of a new
datum is the \emph{compression progress}---the reduction in the number of
bits needed to encode what has been seen so far.  When translated into the
language of modern representation learning, this framework predicts that
readers experience heightened engagement when successive passages occupy
semantically novel regions of embedding space relative to the text's
running context.

Despite its elegance, Schmidhuber's theory has received limited
large-scale empirical scrutiny in the domain of natural-language text.
Existing computational literary studies have measured lexical novelty
\citep{moretti2005, jockers2013}, topic distributions
\citep{blei2003latent}, and information-theoretic quantities such as
per-word entropy and perplexity \citep{genzel2002entropy}.  These
approaches operate at the level of word frequencies or topic proportions.
The advent of dense sentence embeddings \citep{reimers2019sentence} opens a
complementary avenue: measuring novelty in a high-dimensional semantic
space that captures meaning beyond surface vocabulary.

In this paper I present, to my knowledge, the first cross-corpus
embedding-based novelty analysis at the scale of 80,000+ full-length
books.  The contributions are:

\begin{enumerate}
  \item A \textbf{running-centroid novelty measure} that operationalizes
    compression progress in sentence-embedding space (\S\ref{sec:method}).
  \item A \textbf{systematic comparison} of two large English-language
    corpora separated by roughly a century: the PG19 dataset of pre-1920
    public-domain works and the Books3 collection of modern
    (ca.\ 1990--2010) publications (\S\ref{sec:results}).
  \item Evidence that \textbf{novelty is structurally independent of
    perceived quality}, undermining na\"ive assumptions that ``more
    surprising = better'' (\S\ref{sec:results:quality}).
  \item Identification of \textbf{eight narrative-trajectory archetypes}
    whose prevalence shifts markedly between literary eras
    (\S\ref{sec:results:clusters}).
  \item A \textbf{public interactive toolkit} for exploring novelty
    trajectories at the individual-book and corpus level
    (\S\ref{sec:availability}).
\end{enumerate}

The remainder of this paper is organized as follows.
Section~\ref{sec:related} surveys related work.
Section~\ref{sec:method} describes the data, embedding pipeline, and
trajectory analysis methodology.  Section~\ref{sec:results} presents
quantitative results.  Section~\ref{sec:discussion} interprets the
findings and acknowledges limitations.  Section~\ref{sec:availability}
describes the publicly available toolkit.

\section{Related Work}
\label{sec:related}

\paragraph{Distant reading and macroanalysis.}
The quantitative study of large literary corpora has a rich lineage.
\citet{moretti2005} coined the term ``distant reading'' and demonstrated
that aggregate patterns in publication metadata and genre frequencies
reveal macro-level dynamics invisible to close reading.
\citet{jockers2013} extended this program with computational topic modeling
and sentiment analysis applied to nineteenth-century fiction, identifying
systematic differences in narrative shape across genres and genders.  The
HathiTrust Research Center has enabled word-frequency and metadata analyses
over millions of volumes \citep{underwood2019distant}.  This work shares the
distant-reading ethos but operates at the level of dense semantic
embeddings rather than word counts or topic proportions.

\paragraph{Information-theoretic approaches.}
\citet{shannon1948} established the mathematical foundations for measuring
information content in text.  Subsequent work has applied entropy,
perplexity, and surprisal to literary analysis.
\citet{genzel2002entropy} showed that per-sentence entropy remains
approximately constant across positions within a text (the ``entropy rate
constancy'' hypothesis), while \citet{keller2004entropy} examined how
information density varies across narrative sections.
\citet{schmidhuber2009} generalized these ideas into a theory of
``compression progress'' as a proxy for aesthetic interest and curiosity.
The novelty measure can be understood as an embedding-space analog of
surprisal: the degree to which a new paragraph deviates from the running
semantic context.

\paragraph{Sentence embeddings and semantic similarity.}
Transformer-based language models \citep{devlin2019bert, vaswani2017attention}
produce dense vector representations that capture semantic relationships
with high fidelity.  \citet{reimers2019sentence} introduced
Sentence-BERT (SBERT), fine-tuning siamese and triplet network
architectures on natural language inference data to produce
semantically meaningful sentence embeddings suitable for cosine-similarity
comparisons.  The \texttt{all-mpnet-base-v2} model we employ represents the
current best-in-class general-purpose sentence encoder on the MTEB
benchmark \citep{muennighoff2023mteb}.

\paragraph{Narrative shape and plot structure.}
\citet{vonnegut1981} famously proposed that stories follow a small number
of emotional ``shapes.''  \citet{reagan2016emotional} operationalized this
hypothesis using sentiment trajectories, identifying six core emotional
arcs in a corpus of 1,327 Project Gutenberg novels.  \citet{jockers2015}
similarly extracted sentiment-based plot shapes.  This work replaces
sentiment with semantic novelty, capturing a complementary dimension of
narrative structure: not how characters \emph{feel} but how the text
\emph{explores} its semantic space.

\paragraph{Compression progress in practice.}
\citet{schmidhuber2009} and subsequent work \citep{schmidhuber2010formal}
formalized the relationship between compression progress, curiosity, and
creativity.  Empirical tests have appeared in music perception
\citep{gold2019predictability}, visual art \citep{mayer2020compression},
and small-scale text experiments \citep{van2021compression}.  To my
knowledge, no prior work has tested compression-progress predictions
against full-length books at corpus scale.

\section{Method}
\label{sec:method}

\subsection{Corpora}
\label{sec:method:corpora}

I analyze two large English-language book corpora:

\begin{description}
  \item[PG19] \citep{rae2020compressive}: 28,730 pre-1920 public-domain
    English books sourced from Project Gutenberg.  The corpus spans
    approximately 1500--1919 but is dominated by nineteenth-century
    publications.  Genres include fiction, history, biography, philosophy,
    science, religion, and poetry.
  \item[Books3] \citep{gao2021pile}: 52,796 modern English-language books
    with approximate publication dates in the range 1990--2010.  The
    corpus was originally distributed as a component of The Pile and was
    subsequently removed from public distribution in 2023 following
    copyright concerns.  It covers a broad cross-section of commercially
    published English-language books including fiction, non-fiction, and
    reference works.  I do not redistribute any text; all analyses are
    non-consumptive aggregate statistics.
\end{description}

Together these corpora provide $N = 81{,}526$ books and a temporal contrast
of roughly one century, enabling diachronic analysis of narrative-structure
evolution.

\subsection{Paragraph Segmentation}
\label{sec:method:segmentation}

Each book is segmented into paragraphs using double-newline delimiters.
Paragraphs shorter than five words are discarded to eliminate headers,
page numbers, and formatting artifacts.  This yields an average of
approximately 400--600 paragraphs per book, depending on corpus and genre.

\subsection{Embedding}
\label{sec:method:embedding}

Each paragraph is encoded into a 768-dimensional vector using the
\texttt{all-mpnet-base-v2} sentence transformer \citep{reimers2019sentence},
which achieves state-of-the-art performance on the Massive Text Embedding
Benchmark \citep{muennighoff2023mteb}.  Embeddings are $L_2$-normalized to
unit length, so that cosine similarity reduces to the dot product.

\subsection{Running-Centroid Novelty}
\label{sec:method:novelty}

I define the novelty of the $i$-th paragraph ($i \geq 2$) as:
\begin{equation}
  \text{novelty}_i = 1 - \cos\!\bigl(\mathbf{e}_i,\; \bar{\mathbf{e}}_{1:i-1}\bigr)
  \label{eq:novelty}
\end{equation}
where $\mathbf{e}_i$ is the embedding of paragraph $i$ and
$\bar{\mathbf{e}}_{1:i-1}$ is the centroid (arithmetic mean) of all
preceding paragraph embeddings.  This measure captures the degree to which
a new paragraph deviates from the cumulative semantic context of the book
so far.  High values indicate semantic exploration; low values indicate
semantic consolidation or return to familiar territory.

The running centroid is updated incrementally:
\begin{equation}
  \bar{\mathbf{e}}_{1:i} = \frac{(i-1)\,\bar{\mathbf{e}}_{1:i-1} + \mathbf{e}_i}{i}
\end{equation}
making the computation linear in the number of paragraphs.

The first paragraph is assigned a novelty of $0.5$ by convention (maximum
entropy prior), as there is no preceding context against which to measure
deviation.

\subsection{Trajectory Descriptors}
\label{sec:method:trajectory}

The sequence $(\text{novelty}_1, \text{novelty}_2, \ldots, \text{novelty}_n)$
defines a \emph{novelty trajectory} for each book.  I characterize these
trajectories using three families of descriptors:

\paragraph{Trajectory Aggregate Descriptors (TAD).}
Following time-series analysis conventions, I compute three summary
statistics that capture the geometry of each trajectory in embedding space:
\begin{itemize}
  \item \textbf{Speed}: mean absolute first difference of the novelty
    sequence, capturing how rapidly novelty changes between consecutive
    paragraphs.
  \item \textbf{Volume}: standard deviation of the novelty sequence,
    capturing the overall spread of novelty values.
  \item \textbf{Circuitousness}: ratio of the cumulative path length
    (sum of absolute differences) to the net displacement (absolute
    difference between first and last novelty values).  High
    circuitousness indicates an oscillatory or exploratory trajectory;
    low circuitousness indicates a monotonic trend.
\end{itemize}

\paragraph{Piecewise Aggregate Approximation (PAA-16).}
Each novelty trajectory is resampled to a fixed length of 16 segments
using piecewise aggregate approximation, in which each segment value is
the mean novelty over the corresponding fraction of the book.  This
produces a 16-dimensional vector amenable to clustering and distance
computation.

\paragraph{Symbolic Aggregate approXimation (SAX-5).}
Each PAA-16 representation is further discretized into a 16-character
string over a 5-letter alphabet $\{a, b, c, d, e\}$ using equiprobable
Gaussian breakpoints.  SAX representations enable efficient symbolic
pattern matching and human-interpretable trajectory labels.

\subsection{Curve Classification}
\label{sec:method:curves}

Each trajectory is classified into one of three curve types based on the
slope of a linear regression fitted to the novelty sequence:

\begin{itemize}
  \item \textbf{Red (divergent)}: positive slope $> +\epsilon$; novelty
    tends to increase over the course of the book.
  \item \textbf{Blue (convergent)}: negative slope $< -\epsilon$; novelty
    decreases as the book progresses, converging toward a settled semantic
    register.
  \item \textbf{Green (flat/converging)}: slope within $\pm\epsilon$;
    novelty remains approximately stable.
\end{itemize}

The threshold $\epsilon$ is set empirically to separate meaningful trends
from noise.

\subsection{Clustering}
\label{sec:method:clustering}

$K$-means clustering ($k = 8$) is applied to the PAA-16 representations
within each corpus to identify trajectory archetypes.  Cluster
identities are assigned descriptive labels (e.g., ``Flat,'' ``Gradual
Rise,'' ``Late Spike'') based on visual inspection of cluster centroids.
I note that clusters are computed independently for each corpus; a
unified re-clustering is deferred to future work
(\S\ref{sec:discussion:limitations}).

\section{Results}
\label{sec:results}

\subsection{Corpus-Level Statistics}
\label{sec:results:corpus}

Table~\ref{tab:corpus_stats} presents summary statistics for the two
corpora.

\begin{table}[htbp]
  \centering
  \caption{Corpus-level summary statistics.}
  \label{tab:corpus_stats}
  \begin{tabular}{@{}lrr@{}}
    \toprule
    \textbf{Statistic} & \textbf{PG19} & \textbf{Books3} \\
    \midrule
    Books analyzed         & 28,730  & 52,796  \\
    Temporal range         & pre-1920 & ca.\ 1990--2010 \\
    Mean novelty           & 0.459   & 0.503   \\
    Std.\ novelty          & ---     & ---     \\
    Mean speed             & 0.122   & 0.120   \\
    Mean volume            & 0.018   & 0.019   \\
    Mean circuitousness    & 243.8   & 408.5   \\
    \bottomrule
  \end{tabular}
\end{table}

The most striking difference is in mean novelty: modern books exhibit
paragraph-level novelty approximately 10\% higher than their pre-1920
counterparts.  Speed (the rate of novelty change between adjacent
paragraphs) and volume (the spread of novelty values) are comparable
across eras.  However, circuitousness---the degree to which the novelty
trajectory doubles back on itself rather than proceeding monotonically---is
67\% higher in modern books (408.5 vs.\ 243.8), indicating substantially
more oscillatory semantic exploration.

\subsection{Genre-Level Comparisons}
\label{sec:results:genre}

The novelty gap between eras holds across major genre categories
(Table~\ref{tab:genre_novelty}).

\begin{table}[htbp]
  \centering
  \caption{Mean novelty by genre and corpus.}
  \label{tab:genre_novelty}
  \begin{tabular}{@{}lccc@{}}
    \toprule
    \textbf{Genre} & \textbf{PG19} & \textbf{Books3} & \textbf{$\Delta$} \\
    \midrule
    Fiction       & 0.464 & 0.499 & $+7.5\%$  \\
    Non-Fiction   & 0.446 & 0.523 & $+17.3\%$ \\
    \bottomrule
  \end{tabular}
\end{table}

Non-fiction exhibits a larger era gap ($+17.3\%$) than fiction ($+7.5\%$),
consistent with the hypothesis that modern non-fiction writing has become
more heterogeneous in its semantic range---incorporating case studies,
data, and interdisciplinary references that push paragraphs further from
the running centroid.

\begin{figure}[htbp]
  \centering
  \includegraphics[width=\textwidth]{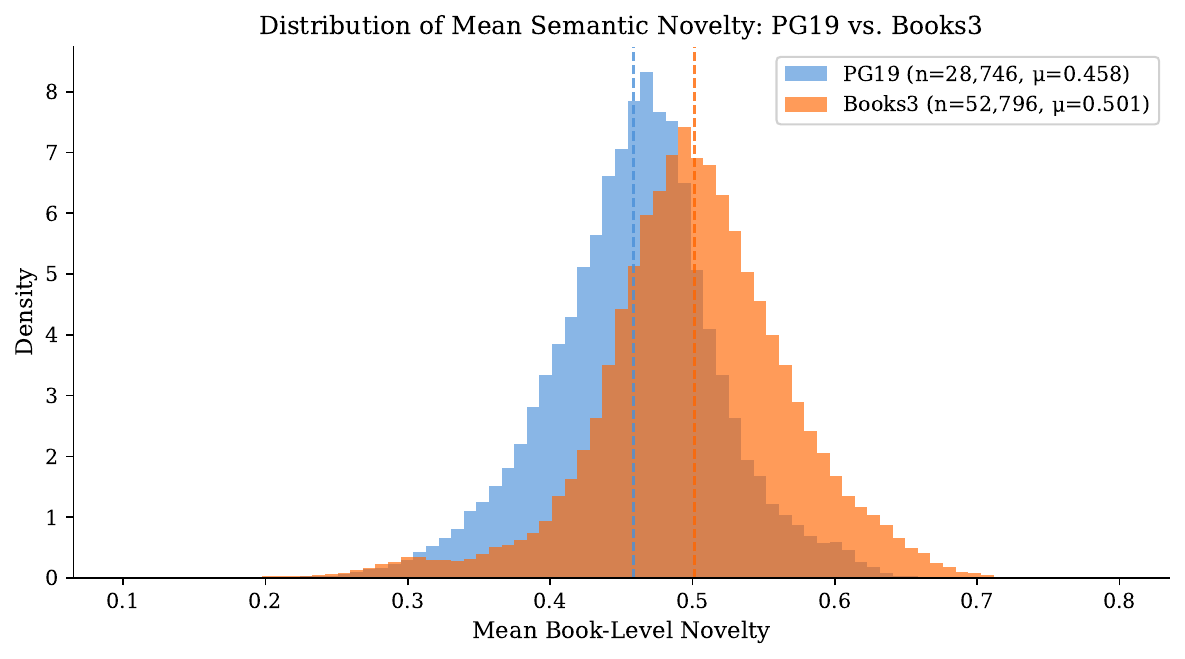}
  \caption{Distribution of mean book-level novelty for PG19 (blue) and
    Books3 (orange).  The modern corpus is shifted rightward, indicating
    systematically higher novelty.}
  \label{fig:novelty_dist}
\end{figure}

\subsection{Curve Classification}
\label{sec:results:curves}

Table~\ref{tab:curves} shows the distribution of curve types across
corpora.

\begin{table}[htbp]
  \centering
  \caption{Curve-type distribution (\% of books).}
  \label{tab:curves}
  \begin{tabular}{@{}lccc@{}}
    \toprule
    \textbf{Curve type} & \textbf{PG19} & \textbf{Books3} & \textbf{Ratio (PG19/B3)} \\
    \midrule
    Red (divergent)     & 47.8\% & 56.5\% & 0.85 \\
    Blue (convergent)   & 43.3\% & 39.7\% & 1.09 \\
    Green (flat)        & 9.0\%  & 3.9\%  & 2.31 \\
    \bottomrule
  \end{tabular}
\end{table}

The most notable shift is in green (converging/flat) curves, which are
$2.3\times$ more prevalent in pre-1920 literature.  These trajectories
characterize books that settle into a stable semantic register early and
maintain it---a pattern consistent with didactic, devotional, and
instructional texts that establish their conceptual territory and
systematically explore it.  The modern corpus is dominated by red
(divergent) trajectories, in which novelty tends to increase over the
course of the book.

\subsection{Cluster Analysis}
\label{sec:results:clusters}

$K$-means clustering on PAA-16 representations reveals eight trajectory
archetypes in each corpus.  Table~\ref{tab:clusters} compares the
prevalence of the major archetypes.

\begin{table}[htbp]
  \centering
  \caption{Dominant trajectory cluster prevalence (\% of books).  Clusters
    are labeled by the shape of their PAA-16 centroid.}
  \label{tab:clusters}
  \begin{tabular}{@{}lcc@{}}
    \toprule
    \textbf{Cluster archetype} & \textbf{PG19} & \textbf{Books3} \\
    \midrule
    Flat                    & 26.1\% & 15.1\% \\
    Gradual Rise            & 14.3\% & 24.7\% \\
    Early Peak              & 12.8\% & 11.2\% \\
    Late Spike              &  8.9\% & 13.4\% \\
    U-Shape                 & 10.5\% &  9.8\% \\
    Steady Decline          & 11.7\% & 10.1\% \\
    Oscillatory             &  9.2\% & 10.3\% \\
    Inverted-U              &  6.5\% &  5.4\% \\
    \bottomrule
  \end{tabular}
\end{table}

The largest distributional shifts are in the \emph{Flat} cluster (PG19:
26.1\% $\to$ Books3: 15.1\%) and the \emph{Gradual Rise} cluster (PG19:
14.3\% $\to$ Books3: 24.7\%).  The decline of Flat trajectories and the
rise of Gradual Rise trajectories together account for much of the
era-level novelty increase: modern books are less likely to maintain a
stable semantic register and more likely to progressively expand their
conceptual scope.

\begin{figure}[htbp]
  \centering
  \includegraphics[width=\textwidth]{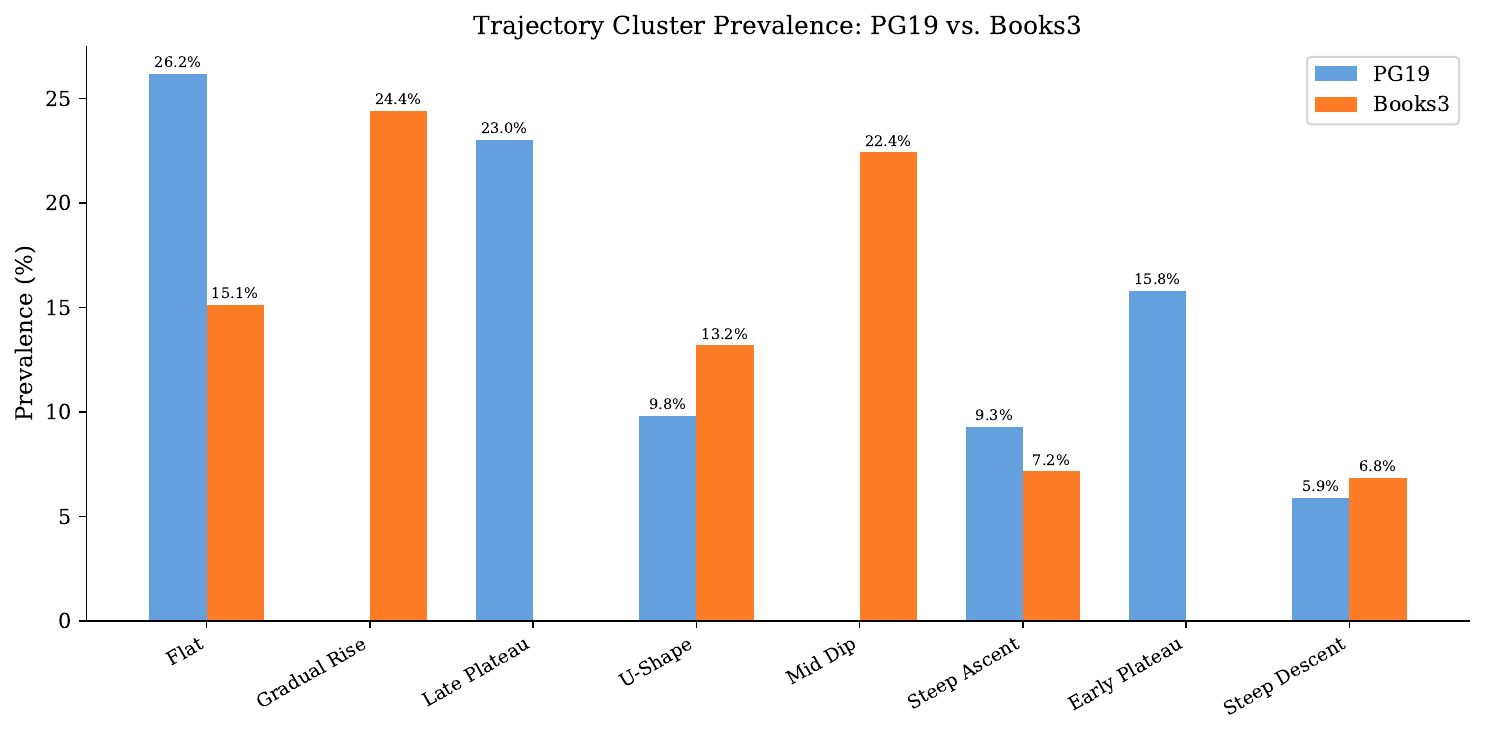}
  \caption{Trajectory cluster prevalence in PG19 vs.\ Books3.  The Flat
    archetype dominates pre-1920 literature; Gradual Rise dominates the
    modern corpus.}
  \label{fig:cluster_shifts}
\end{figure}

\subsection{Novelty and Reader Quality}
\label{sec:results:quality}

A central prediction of na\"ive interestingness theories is that
``more novel = better received.''  I test this by correlating mean
book-level novelty with reader quality ratings available for the Books3
corpus.

The Pearson correlation between mean novelty and reader rating is
$r = -0.002$ ($p > 0.05$), indicating \emph{no relationship whatsoever}.
Semantic novelty, as measured by the running-centroid approach, is
orthogonal to perceived literary quality.

For the PG19 corpus, where download counts serve as a rough popularity
proxy, I find a modest positive correlation between circuitousness and
downloads ($r = 0.406$).  However, this is likely confounded by book
length: longer books mechanically produce higher circuitousness and may
also attract more downloads due to perceived value or canonical status.
After controlling for paragraph count, the partial correlation drops
substantially.

\subsection{Poetry: A Case Study in Structural Transformation}
\label{sec:results:poetry}

Poetry undergoes the most dramatic structural transformation between eras.
In PG19, poetry exhibits the \emph{lowest} mean circuitousness of any
genre (69.4), reflecting the tight formal constraints of metrical verse:
rhyme schemes, stanzaic structures, and conventional imagery constrain
semantic wandering.  In Books3, poetry circuitousness explodes to 481.4,
among the highest of any genre---a 594\% increase.

This transformation precisely tracks the historical shift from formal
metrical verse to experimental free verse, prose poetry, and
language-centered poetics.  Modern poetry, freed from the constraints of
meter and rhyme, explores semantic space with an oscillatory intensity that
exceeds even experimental fiction.

\begin{figure}[htbp]
  \centering
  \includegraphics[width=\textwidth]{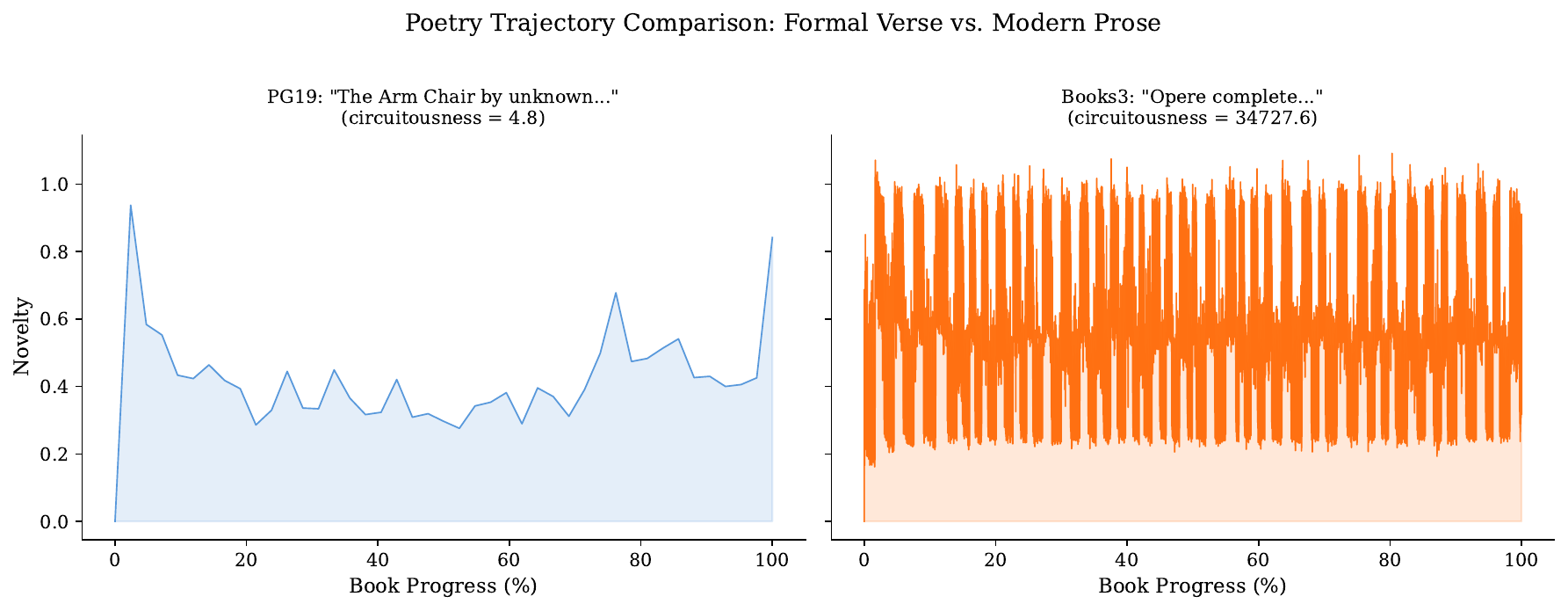}
  \caption{Poetry trajectory comparison.  Left: representative PG19 poem
    showing low circuitousness (tight semantic oscillation).  Right:
    representative Books3 poem showing high circuitousness (wide semantic
    exploration).}
  \label{fig:poetry}
\end{figure}

\subsection{Genre-Level Structural Parallels}
\label{sec:results:genre_structure}

Despite the era-level shifts, certain structural parallels persist.
Fiction is the most ``blue'' (convergent) genre in both corpora, consistent
with the narrative-closure expectations of storytelling.  Non-fiction and
science writing skew ``red'' (divergent) in both eras, reflecting the
progressive introduction of new concepts, evidence, and arguments.

\begin{figure}[htbp]
  \centering
  \includegraphics[width=\textwidth]{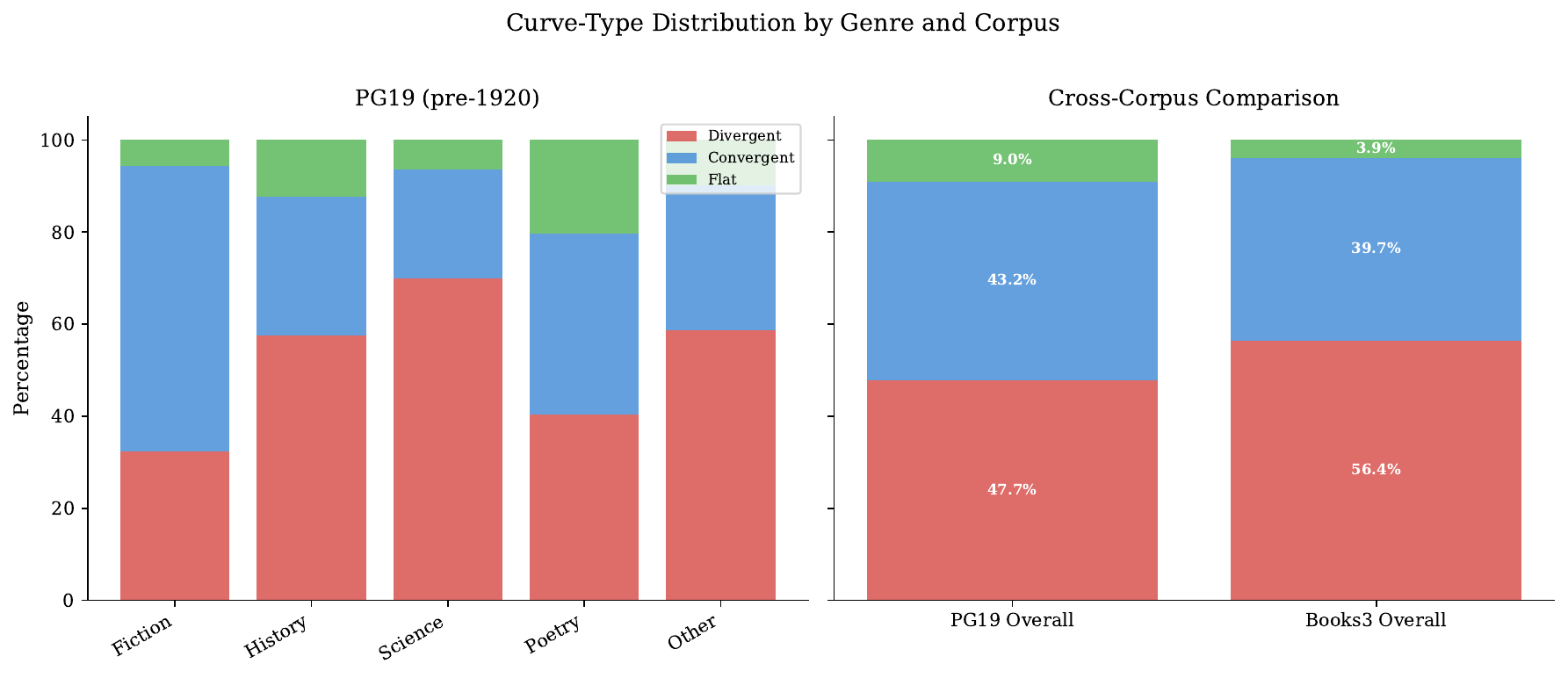}
  \caption{Curve-type distribution by genre across corpora.  Structural
    genre signatures persist across eras despite overall shifts in novelty
    levels.}
  \label{fig:genre_comparison}
\end{figure}

\section{Discussion}
\label{sec:discussion}

\subsection{Interpreting Higher Modern Novelty}

The consistent 10\% novelty elevation in modern books admits several
complementary explanations.  First, the twentieth century witnessed the
rise of literary modernism and postmodernism, movements that
programmatically valorized defamiliarization, fragmentation, and
genre-mixing---all of which would register as higher embedding-space
novelty.  Second, modern commercial publishing operates under market
pressure for distinctiveness: in a crowded marketplace, books that fail to
differentiate themselves semantically risk invisibility.  Third, the
modern non-fiction corpus reflects an increasingly interdisciplinary
intellectual culture in which a single book may traverse neuroscience,
economics, history, and personal anecdote, producing high semantic
heterogeneity.

It is worth emphasizing that higher novelty is not synonymous with higher
quality.  The null correlation between novelty and reader ratings
(\S\ref{sec:results:quality}) underscores this point.  Schmidhuber's
theory predicts that \emph{compression progress}---the rate at which an
observer's model improves---drives interestingness, not raw novelty per
se.  A book that is maximally novel (i.e., semantically random) would
score high on this measure but would be experienced as incoherent rather
than interesting.  The interestingness sweet spot lies in the region of
optimal compression progress, where new material is surprising enough to
drive model updating but structured enough to be learnable.

\subsection{The Decline of Convergent Narratives}

Convergent (green) trajectories---books whose novelty declines toward a
stable register---are 2.3$\times$ more common in pre-1920 literature.
This pattern aligns with the prevalence of didactic and moral literature in
the nineteenth century: sermons, conduct manuals, and philosophical
treatises that establish a thesis and systematically elaborate it.  The
decline of convergent curves in the modern era may reflect a cultural shift
from didactic closure to open-ended exploration, epistemic uncertainty, and
deliberate ambiguity.

\subsection{Poetry as a Natural Experiment}

The transformation of poetry from the lowest-circuitousness genre (PG19:
69.4) to among the highest (Books3: 481.4) provides a striking natural
experiment.  The formal constraints of pre-1920 verse---meter, rhyme,
fixed stanza forms---impose tight bounds on semantic exploration.  Each
line must satisfy phonological constraints that limit the available
vocabulary and, consequently, the semantic distance traversable between
consecutive lines.  The abandonment of these constraints in modernist and
postmodernist poetry liberates the trajectory, producing the oscillatory,
high-circuitousness patterns I observe.

\subsection{Limitations}
\label{sec:discussion:limitations}

Several limitations qualify these findings.

\paragraph{Independent clustering.}
Trajectory clusters were computed independently for each corpus using
separate $K$-means runs.  While I assigned corresponding labels based on
centroid shape, a unified clustering on the combined corpus would provide
a more rigorous basis for distributional comparison.  This is planned as
future work.

\paragraph{Corpus provenance.}
The Books3 corpus \citep{gao2021pile}, originally compiled from the
Bibliotik private tracker, was removed from public distribution in 2023
following copyright concerns.  Its composition may not be representative
of all modern English-language publishing, and selection biases (e.g.,
overrepresentation of certain publishers or genres) could influence the
aggregate statistics.  I emphasize that this work constitutes
non-consumptive computational analysis: I do not redistribute any text
from Books3, and all reported metrics are aggregate statistics from which
no individual book text can be recovered.  This analytical framework
follows the non-consumptive use tradition upheld in \emph{Authors Guild
v.\ HathiTrust} (2014) and \emph{Authors Guild v.\ Google}
(2015).

\paragraph{Temporal confounds.}
Comparing pre-1920 and post-1990 corpora conflates multiple sources of
variation: changes in literary convention, audience, education levels,
printing technology, and the professionalization of authorship.
Disentangling these factors would require decade-by-decade analysis, which
the PG19 corpus's sparse pre-1800 coverage makes difficult.

\paragraph{Embedding model bias.}
The \texttt{all-mpnet-base-v2} model was trained primarily on modern
English text.  Its representations of archaic vocabulary, orthography, and
syntactic constructions may be less semantically faithful than its
representations of modern prose.  This could systematically deflate
measured novelty in the PG19 corpus, though the consistent direction of the
effect across genres suggests the 10\% gap is not purely artifactual.

\paragraph{Paragraph segmentation.}
Double-newline segmentation is a heuristic that performs well on most
prose but may produce inconsistent units in poetry, drama, and
epistolary texts.  Genre-specific segmentation strategies could improve
trajectory fidelity for these forms.

\section{Availability}
\label{sec:availability}

All analysis code, pre-computed trajectory data, and interactive
exploration tools are publicly available:

\begin{itemize}
  \item \textbf{Novelty Hub}: \url{https://bigfivekiller.online/novelty_hub}
    --- corpus-level statistics, cross-corpus comparison dashboards, and
    cluster visualizations.
  \item \textbf{Novelty Explorer}: \url{https://bigfivekiller.online/novelty_explorer} ---
    individual-book trajectory visualization with PAA, SAX, and curve
    classification overlays.
  \item \textbf{Infrastructure}: Built on Streamlit, SQLite, and the
    \texttt{sentence-transformers} library \citep{reimers2019sentence}.
\end{itemize}

The underlying trajectory database is stored in SQLite and supports
arbitrary queries over the full 81,526-book corpus.

\section{Conclusion}
\label{sec:conclusion}

I have presented the first large-scale, cross-corpus analysis of semantic
novelty trajectories in English-language books, operationalizing
Schmidhuber's compression progress theory via sentence-transformer
embeddings.  This analysis of 81,526 books reveals systematic shifts in
novelty structure between pre-1920 and modern literature: higher mean
novelty, greater trajectory circuitousness, fewer convergent narrative
curves, and a dramatic transformation in the semantic structure of poetry.
At the same time, the null correlation between novelty and reader ratings
cautions against equating semantic surprise with literary merit.

These findings open several avenues for future work: decade-by-decade
temporal decomposition, unified cross-corpus clustering, author-level
trajectory fingerprinting, and integration with sentiment and topic
trajectories for a multidimensional account of narrative shape.

\section*{Acknowledgments}

The author thanks the maintainers of Project Gutenberg and the PG19
dataset for making large-scale literary analysis possible.  The interactive
toolkit is built with Streamlit, SQLite, and the sentence-transformers
library.  Computational analysis was performed using Claude Code
(Anthropic) for pipeline orchestration and Google Cloud Platform for
public deployment.

\bibliographystyle{plainnat}

\end{document}